\documentclass{ecai} 

\usepackage{latexsym}
\usepackage{amssymb}
\usepackage{amsmath}
\usepackage{amsthm}
\usepackage{booktabs}
\usepackage{enumitem}
\usepackage{graphicx}
\usepackage{color}
\usepackage{xcolor}

\usepackage{xspace}
\usepackage{multirow}
\usepackage{tabularray}
\usepackage{bbding}
\usepackage{parskip}

\theoremstyle{remark}

\newcommand{\BibTeX}{B\kern-.05em{\sc i\kern-.025em b}\kern-.08em\TeX}

\newcommand{\ie}{{\emph{i.e.}},\xspace}
\newcommand{\eg}{{\emph{e.g.}},\xspace}

\newcommand{\APBEV}{AP$_\text{BEV}$\xspace}
\newcommand{\APthreeD}{AP$_\text{3D}$\xspace}
\newcommand{\APnew}{AP$_\text{CS-BEV}$\xspace}
\newcommand{\APcs}{AP$_\text{CS-ABS}$\xspace}

\begin{document}


\begin{frontmatter}




\title{Detect Closer Surfaces that can be Seen: New Modeling and Evaluation in Cross-domain 3D Object Detection}



\author[A]{\fnms{Ruixiao}~\snm{Zhang}\thanks{Corresponding Author. Email: rz6u20@soton.ac.uk.}}
\author[A]{\fnms{Yihong}~\snm{Wu}}
\author[B]{\fnms{Juheon}~\snm{Lee}\footnote{This study was conducted before joining Meta.}}
\author[A]{\fnms{Adam}~\snm{Prugel-Bennett}}
\author[A]{\fnms{Xiaohao}~\snm{Cai}} 

\address[A]{Department of Electronic and Computer Science, University of Southampton}
\address[B]{Meta Inc.}


\begin{abstract}
The performance of domain adaptation technologies has not yet reached an ideal level in the current 3D object detection field for autonomous driving, which is mainly due to significant differences in the size of vehicles, as well as the environments they operate in when applied across domains. These factors together hinder the effective transfer and application of knowledge learned from specific datasets. Since the existing evaluation metrics are initially designed for evaluation on a single domain by calculating the 2D or 3D overlap between the prediction and ground-truth bounding boxes, they often suffer from the overfitting problem caused by the size differences among datasets. This raises a fundamental question related to the evaluation of the 3D object detection models' cross-domain performance: Do we really need models to maintain excellent performance in their original 3D bounding boxes after being applied across domains? From a practical application perspective, one of our main focuses is actually on preventing collisions between vehicles and other obstacles, especially in cross-domain scenarios where correctly predicting the size of vehicles is much more difficult. In other words, as long as a model can accurately identify the closest surfaces to the ego vehicle, it is sufficient to effectively avoid obstacles. In this paper, we propose two metrics to measure 3D object detection models' ability of detecting the closer surfaces to the sensor on the ego vehicle, which can be used to evaluate their cross-domain performance more comprehensively and reasonably. Furthermore, we propose a refinement head,  named EdgeHead, to guide models to focus more on the learnable closer surfaces, which can greatly improve the cross-domain performance of existing models not only under our new metrics, but even also under the original BEV/3D metrics. Our code is available at {\url{https://github.com/Galaxy-ZRX/EdgeHead}}.
\end{abstract}
\end{frontmatter}


\section{Introduction}
\label{sec:intro}


3D object detection aims to localize and categorize different types of objects in specific 3D space described by 3D sensor data (\eg LiDAR point clouds). Recently, the application of this technology has achieved significant improvement due to the development of deep neural networks, especially in the field of autonomous driving. Current 3D object detection methods mainly focus on specific datasets, \ie models will be trained and tested independently on a specific dataset. In doing so, a number of models achieved high performances on public benchmarks including nuScenes~\cite{9156412}, Waymo~\cite{Sun_2020_CVPR}, and KITTI~\cite{6248074}.  However, if the application of the model on a new dataset is needed, the training on the new dataset as well as modifications of some training hyper-parameters are usually necessary. In other words, it is hard for models trained on one dataset to adapt directly to another. These domain shifts may arise from different sensor types, weather conditions~\cite{9156543}, and object sizes~\cite{9578132} between different datasets or domains. This domain adaptation problem is therefore a big challenge for real-world applications of existing 3D object detection methods, as their retraining steps can be very slow and resource-consuming. It is thus significant to understand the reasons for the drop in cross-domain performance and propose efficient methods to raise the cross-domain performance to the same level as within-domain tasks.

\begin{figure}[t]
\centering
   \includegraphics[width=0.95\linewidth,height=0.62\linewidth]{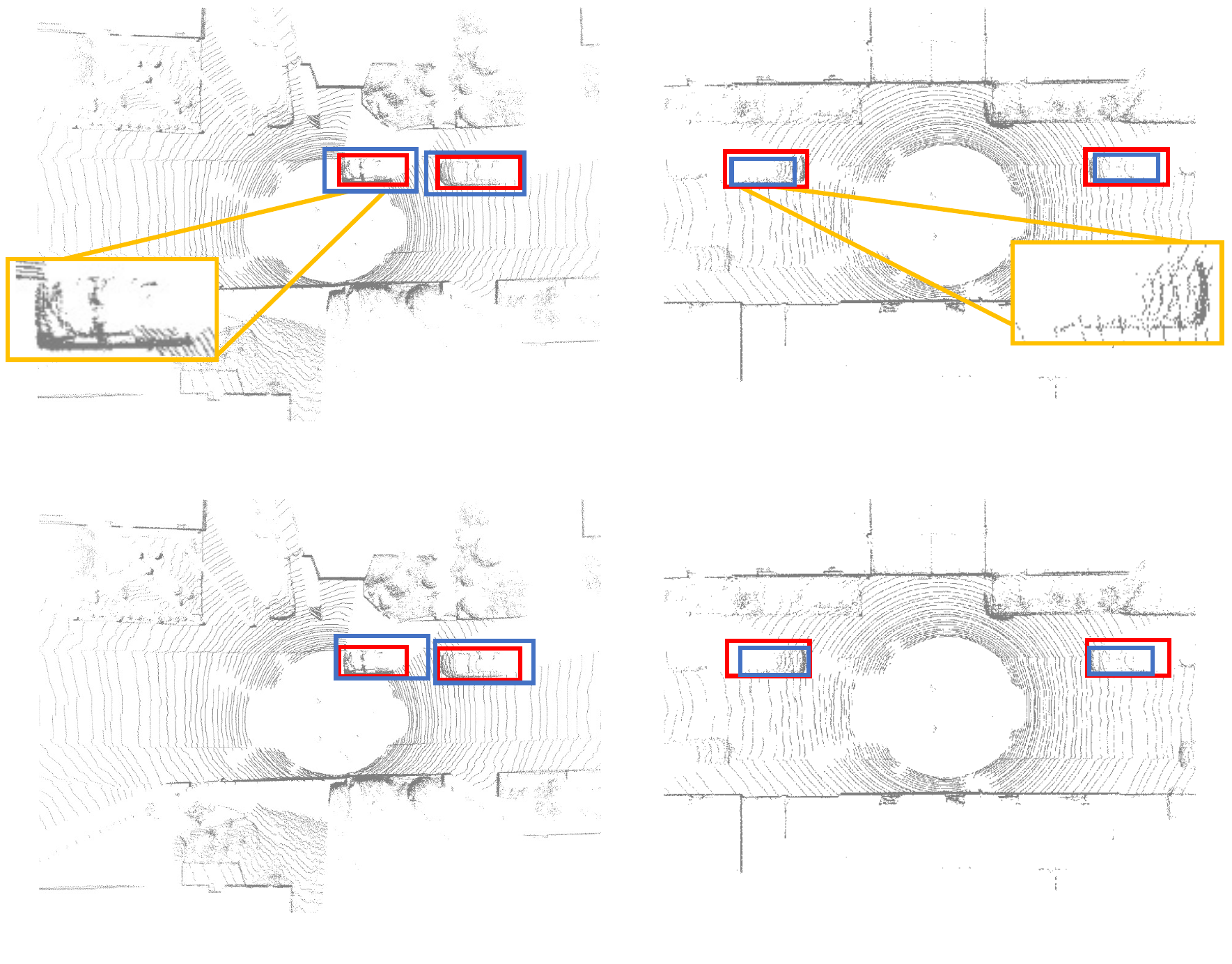}
   \put(-155,83){\footnotesize (a) Without our EdgeHead}
   \put(-150,5){\footnotesize (b) With our EdgeHead}
\vspace{-0.10in}   
\caption{\textbf{Prediction properties of models before and after equipping with our proposed EdgeHead in cross-domain tasks.}  Red: Ground-truth boxes. Blue: Predictions. The left and right columns showcase the prediction properties when the training domain respectively has a larger and smaller average object size than the target domain. Object size overfitting problem occurs in both cases, while our proposed EdgeHead can help better detect the closer surfaces to the ego vehicle.}
\label{fig:first}
\end{figure}


By comparing the performance of two models~\cite{8954080, 8578896} on several different datasets, the work in \cite{9156543} showed that the difference in car size across geographic locations is one of the main challenges for domain adaptation problems. 
Other factors such as the difference in LiDAR point cloud densities~\cite{wei2022lidar, hu2023density} and weather conditions~\cite{9156543} have also been investigated.


Some generic training methods have tried to overcome the domain adaptation problem. Among these, some methods explore the possibility of adding information about domain gaps into the models by training or using additional prior knowledge~\cite{9156543, 9578132, yuan2023bi3d, chen2023revisiting} to adapt the trained models from the training domain to new domains. Some other methods tried to improve the models' generalization ability to enhance their performance on multiple domains without fine-tuning and additional training operations~\cite{hu2023density, zhang2023uni3d}. It is common practice for methods to evaluate their cross-domain performance using established 3D object detection metrics designed for single-domain tasks, such as bird’s-eye view average precision (BEV AP) and 3D average precision (3D AP). Since these metrics are initially designed for the performance measurement on a single domain, they do not consider the differences across domains and therefore easily suffer from the overfitting problem in object size, making them difficult to evaluate models' cross-domain performance comprehensively.

From another point of view, existing metrics are usually designed to evaluate models' ability to predict the complete shapes of objects. However, as shown in Figure~\ref{fig:first}, most objects' point cloud is incomplete due to physical obstructions in the capturing process of LiDAR sensors. As a result, it is difficult to predict the full size and entire box of objects. In within-domain tasks, we may be able to "guess" the prediction through training with large amounts of samples. However, the guessing easily becomes incorrect in cross-domain tasks since the average object size has changed. Therefore, correctly predicting the location (usually represented by the center of the box) along with the full box size becomes a much more difficult task. Instead, we argue that what we can better guide the model to learn from the incomplete object point cloud is to predict the closer surfaces of the objects to the ego vehicle, since there are plenty of points over there as shown in Figure~\ref{fig:first}. Among the four surfaces perpendicular to the ground, capturing and describing these two is more reasonable and also more important, since our main purpose is to avoid potential collisions with surrounding objects during driving.


In this work, we approach the problem of cross-domain 3D object detection from a different perspective. Our main contributions are as follows: We first propose two novel evaluation metrics that aim to evaluate models' ability to detect the closer surfaces to the sensor. By using the defined absolute gap of the closer surfaces (\ie $G_{cs}$) to penalize the original BEV AP, we propose the {\it closer-surfaces penalized BEV} (\ie the CS-BEV) AP and achieve a balance between the detection quality of the entire box and the closer surfaces to the ego vehicle. Additionally, to avoid the interference of different BEV AP, we introduce the {\it absolute closer-surfaces} (\ie the CS-ABS) AP metric for a more precise analysis of improvements in closer-surfaces detection contributed by different methods.
Furthermore, we design a second-stage refinement head, \ie \textit{EdgeHead}, specifically for the closer-surfaces detection task. It can directly be combined with various existing models. Our EdgeHead can guide models to make predictions with a smaller gap between the closer surfaces and the closest vertex to the ego vehicle.  To the best of our knowledge, this is the first work to improve cross-domain performance with minimal modifications to existing models by adding a novel refinement module.

Thorough experiments show that our proposed metrics can find a balance between evaluating the detection quality of the entire boxes and the closer surfaces to the ego vehicle, and provide a brand new way to describe and measure models' remaining detection ability after adapting to new domains. We also experimentally prove that by simply equipping existing models with the proposed EdgeHead, their detection ability on the closer surfaces can be greatly improved, which not only leads to higher performance under the new proposed metrics but also improves the performance under the original BEV and 3D metrics. The results indicate that by guiding the models to focus more on the unobstructed closer surfaces that can be scanned and surrounded by more points, our proposed strategies can help the models learn more robustly from the point cloud data, and therefore predict both the closer surfaces and the entire boxes better.

In brief, our findings suggest that 3D vision and autonomous driving researchers pay more attention to the detection quality of the objects' closer surfaces to the ego vehicle in cross-domain 3D object detection tasks, which can help models learn more robustly and perform better across different domains.



\section{Related Work}
\label{sec:related}

\noindent\textbf{3D object detection with point clouds.} A common way of representing real-world 3D space is using LiDAR point cloud data, i.e., using 3D points to record the 3D environment. Although it benefits from accurate point locations, the main challenge for LiDAR-based 3D object detection methods is finding the best way to process the point cloud data. CNNs have been widely used in 2D object detection problems; however, due to the sparsity and spatial disorder of the point cloud data, they cannot be directly fed into the CNNs. Therefore, current LiDAR-based methods either transform the point clouds into spatial invariant formats for CNN models or learn features directly from the geometry of 3D points. For example, VoxelNet~\cite{8578570} and SECOND~\cite{s18103337} encoded point clouds into voxels so that features can be extracted by CNNs designed for 3D inputs. MV3D~\cite{8100174} projected point clouds into 2D spaces (\ie the front view and bird's-eye view) and then used 2D CNNs to extract features. PointRCNN~\cite{8954080} applied PointNet++~\cite{NIPS2017_d8bf84be} to obtain 3D point-wise features and directly learned the 3D proposals from the points. PV-RCNN~\cite{Shi_2020_CVPR} voxelized the point cloud data first and then used key-point-wise features to keep more semantic features, in which the combination of point-based and voxel-based methods greatly improved the performance with acceptable computation cost. More recently, some methods~\cite{chen2023voxenext, li2023pillarnext} further explored the potential of convolution-based models and have achieved state-of-the-art performance. With the development of transformers~\cite{dosovitskiy2020vit, carion2020end} in the computer vision field, there are also methods attempting to use transformer-based models to predict 3D objects. For example, TransFusion~\cite{Bai_2022_CVPR} used the transformer decoder to combine the image features and LiDAR point cloud features for  prediction quality enhancement.


\noindent\textbf{Domain adaptation.} Domain adaptation has been widely used in 2D object detection~\cite{Hsu_2020_WACV, 9008383, 8953674} and 2D semantic segmentation~\cite{8578921, pmlr-v80-hoffman18a, Huang_2018_ECCV}. However, there are limited approaches specifically designed for 3D object detection with domain adaptation. \cite{9156543} first proposed the overfitting problem in object sizes and provided a simple but effective solution by normalizing the object size of different datasets based on additional prior knowledge. ST3D~\cite{9578132} and ST3D++~\cite{yang2021st3d++} proposed the random object scaling (ROS) algorithm to solve the overfitting problem in car size and used self-training algorithms to generate pseudo labels and train models on the target domain without ground truth. \cite{wei2022lidar} proposed a distillation method for LiDAR point clouds in order to overcome the beam difference between datasets, which enables models to adapt to point cloud data with lower densities and fewer beams.

\noindent\textbf{Model generalization.} Besides the adaptation methods described above, there are some explorations on improving models' generalization ability, \ie training models only once and enabling them to achieve acceptable performance on more domains. Uni3D~\cite{zhang2023uni3d} aligned the unavoidable differences between datasets via a data-level correction operation and a semantic-level coupling-and-recoupling module. \cite{wu2023towards} proposed a multi-domain knowledge transfer framework to leverage spatial-wise and channel-wise knowledge across domains, which helps extract universal feature representations for models. \cite{hu2023density} proposed the random beam re-sampling (RBRS) method to improve the models' beam-density robustness and used a teacher-student framework to generate pseudo labels on unseen target domains.

\section{Method}
\noindent\textbf{Motivation.}
We first point out the weakness of existing metrics such as BEV and 3D AP in cross-domain tasks. Given two predictions of the same car ground truth as shown in Figure~\ref{fig:first}, exactly the same BEV and 3D AP will be obtained as they have the same overlaps with the ground-truth bounding box. However, there is a big difference when comparing their detection quality of the surfaces that are closer to the LiDAR sensor, which are not occluded by other surfaces and therefore have a bigger chance of being captured by the sensor. This detection quality indicates how correctly a model can estimate the distance from our car to the surfaces of other objects that could collide with us, which is directly related to driving safety and should be paid more attention to than the other two surfaces of detected objects. It is therefore essential to develop new metrics that can accurately assess this detection quality of models, thereby enabling a more reasonable and comprehensive evaluation of performance, particularly for cross-domain tasks.


From another point of view, existing models are easy to overfit on the training domain, on which they are designed and trained to perform well. This limits their detection ability across domains. One of the main factors that causes this overfitting problem is that these models usually have a regression module to learn the offsets of box dimensions and locations between the prediction and ground truth. Such a module results in the overfitting on object sizes, especially for anchor-based methods. It is thus critical to explore the performance of a specifically designed model with the consistent aim of the new metric we usher in here, \ie the one that focuses more on the detection quality of the closer surfaces of the bounding box. 


The aim of our endeavor is therefore twofold. Firstly, we aim to measure models' cross-domain 3D object detection performance with fairer and more reasonable metrics. Specifically, we will design new evaluation metrics that will be less influenced by the cross-domain factors (\eg object sizes and point cloud densities). Secondly, we attempt to improve the models' detection ability that can be preserved across different domains by guiding them with the newly proposed metrics. We will achieve this by equipping existing models with an additional proposed refinement head, called EdgeHead, together with novel proposed loss functions.

\subsection{Problem statement}
The purpose of 3D object detection is to predict the 3D bounding boxes of objects that are parameterized by the center locations $(x_c, y_c, z_c)$, the sizes $(l,w,h)$ and the rotation angle $\theta$. In cross-domain tasks, we train models on the source domain {$\left\{ (X_{i}^{s}, Y_{i}^{s}) \right\} _{i=1}^{N_{s}}$, but focus on their performance on the target domain $\left\{ X_{i}^{t} \right\} ^{N_t}_{i=1}$, where $N_{s}$ and $N_{t}$ are respectively the number of samples in the source domain and target domain, $X_i^s$ and 
$Y_i^s$ respectively denote the $i$-th source domain point cloud data and its corresponding label, and $X_i^t$ denotes the $i$-th target domain point cloud data.

\subsection{Closer-surfaces-based evaluation metrics}\label{sec:metricdesign}
We now propose new evaluation metrics to measure the models' ability to detect the closest corner of an object and its two surfaces that are closer to the LiDAR sensor. 


We first define the absolute gap between the closer surfaces of predictions and ground truth. Given a prediction box with its vertices $\left\{V_{\rm pred}^i \right\} _{i=1}^4$ on the BEV plane and the related ground-truth box with its vertices $\left\{V_{\rm gt}^i \right\} _{i=1}^4$, we first sort their vertices by their distance to the origin (\ie the location of the LiDAR sensor), and then further sort the second and third vertices by their absolute x-coordinate. After sorting, prediction and ground truth boxes should follow the same indexing rule for their vertices, \ie $V^1$ and $V^4$ are respectively the vertices closest and furthest to the origin, and $V^2$ is the vertex having a smaller absolute x-coordinate compared with $V^3$. We can then define the absolute gap, say $G_{\rm cs}$, of the closer surfaces between the prediction and the ground truth, \ie
\begin{align} \label{eq:g-cs}
	G_{\rm cs} = | V_{\rm pred}^1 - V_{\rm gt}^1| \!+\! {\rm Dist}(V_{\rm pred}^2, E_{\rm gt}^{1,2}) \!+\! {\rm Dist}(V_{\rm pred}^3, E_{\rm gt}^{1,3}),
\end{align}
where $E^{i,j}$ is the edge connecting $V^i$ and $V^j$, and ${\rm Dist}(V,E)$ calculates the perpendicular distance from vertex $V$ to edge $E$. 



The defined absolute gap $G_{\rm cs}$ in Eq. \eqref{eq:g-cs} can be used to measure the detection quality of the closer surfaces; however, it will fluctuate with the sizes of the object boxes. In other words, it is not a scaled metric that can be used to calculate the AP with pre-determined thresholds. 
To solve this problem, we propose the \textbf{\textit{absolute closer-surfaces AP}} (\ie the CS-ABS AP) to directly measure the detection quality of the closer surfaces by using
\begin{align} \label{eq:abs-cs}
    \Gamma_{\rm ABS}^{\rm CS} = {1}/{(1+\alpha  G_{\rm cs})},
\end{align}
where $\alpha\ge 0$ is the penalty ratio set to 1 by default. 

The proposed CS-ABS AP by Eq. \eqref{eq:abs-cs} can also be utilized to combine with existing popular metrics and thus form new metrics with hybrid effectiveness for more powerful and fairer evaluation of models' performance.
In particular, we combine the CS-ABS AP with the BEV AP and propose the \textbf{\textit{closer-surfaces penalized BEV AP}} (\ie the CS-BEV AP) to measure the detection quality by using the penalized IoU, say $\Gamma_{\rm BEV}^{\rm CS}$, \ie
\begin{align} \label{eq:bev-cs}
    \Gamma_{\rm BEV}^{\rm CS }= {\Gamma_{\rm BEV}}/{(1+\alpha  G_{\rm cs})}
\end{align}
where $\Gamma_{\rm BEV}$ is the original BEV IoU and $\alpha$ is the penalty ratio set to 1 by default based on experimental experience (see more discussions in the Supplementary Material). 
%
Our proposed CS-BEV metric in Eq. \eqref{eq:bev-cs} not only retains the robustness of the original BEV metric but also better distinguishes the detection quality of the closer surfaces. It finds an evaluation balance between the quality of the entire 3D box and the quality of the closer surfaces. Taking the same examples in Figure~\ref{fig:first}, the newly proposed metric will return a higher AP when the prediction matches the closer surfaces of the ground truth better.



In sum, we in this section proposed two evaluation metrics: CS-ABS AP and CS-BEV AP. The CS-ABS AP can directly tell the detection quality of the closer surfaces without considering the ability to evaluate the quality of the entire 3D box, which can be specifically used when analyzing the detection quality gain regarding closer surfaces. The CS-BEV AP 
can find a balance between the quality of the entire 3D box and the closer surfaces, which is more comprehensive and can be used to measure the overall cross-domain performance for different models and tasks.


\subsection{EdgeHead for closer-surfaces localization}\label{sec:edgehead}
To improve models' closer-surfaces localization ability, we propose a refinement head, \ie the \textbf{\textit{EdgeHead}}, by modifying the models' training purpose. Similar to other refinement heads, the proposed EdgeHead takes the predictions of a model's first stage as the regions of interest (RoIs). It then aggregates the features from earlier backbones of the model (\eg the 3D convolution backbones) into the RoI features for prediction refinement. During the refinement process of EdgeHead, we  modify the loss function to guide the model to learn the closer-surfaces offsets between the predictions and ground truth. 

The voxel RoI pooling~\cite{deng2020voxel} is used to aggregate the RoI features. In detail, we extract the 3D voxel features from the last two layers in the 3D sparse convolution backbone, which is available for most voxel-based 3D object detection models. Afterwards, the feature of each RoI is assigned by aggregating the 3D features from its neighbor voxels via the voxel query operation. Since features from the 3D backbones usually contain more spatial and structural information, the aggregated RoI features can help improve the detection quality of the closer surfaces of bounding boxes. 

The loss of a typical RoI refinement module consists of two parts, \ie the IoU-based classification loss~\cite{Shi_2020_CVPR} and the regression loss. In detail, the original regression loss, say ${\cal L}_{\rm reg}$, uses the smooth $\ell_1$ loss~\cite{7410526} to learn the 7 parameters of the bounding boxes, \ie
\begin{align}\label{eq:origin_loss}
    {\cal L}_{\rm reg} = \sum\limits_{r \in \left\{ x_{\rm c}, y_{\rm c}, z_{\rm c}, l, h, w, \theta \right\} } {\cal L}_{{\rm smooth}-\ell_1}(\widehat{\Delta r^{a}}, \Delta r^{a})
\end{align}
where $\widehat{\Delta r^{a}}$ and $\Delta r^{a}$ are the predicted residual and the regression target, respectively. In our EdgeHead, we use the original classification loss and modify the regression loss in Eq.~\eqref{eq:origin_loss}, which will guide the model to learn the closer-surfaces offsets between the predictions and ground truth.

Given the closest vertex of the anchor box and the ground-truth box respectively as $(x_{\rm cv}^{a}, y_{\rm cv}^{a}, z_{\rm cv}^{a})$ and $(x_{\rm cv}^{\rm gt}$, $y_{\rm cv}^{\rm gt}, z_{\rm cv}^{\rm gt})$, we first rotate the anchor box by the rotation angle $\theta_{\rm gt}$ of the ground-truth box as shown in Figure~\ref{fig:regress_corner}(c), and denote the rotated box's closest vertex by $(x_{\rm cv}^{a^{\prime}}, y_{\rm cv}^{a^{\prime}}, z_{\rm cv}^{a^{\prime}})$. We then calculate the residuals of $x_{\rm cv}$ and $y_{\rm cv}$ between the rotated anchor box and ground truth as follows
\begin{equation}\label{eq:final_reg_loss}
\begin{aligned}
    {\Delta x_{\rm cv}} = x_{\rm cv}^{\rm gt} - x_{\rm cv}^{a^{\prime}}, \quad
    {\Delta y_{\rm cv}} = y_{\rm cv}^{\rm gt} - y_{\rm cv}^{a^{\prime}},
\end{aligned}
\end{equation}
and modify the $\ell_1$ loss by replacing the residual of center locations with the residuals of the rotated closest vertex to the origin (\ie our ego vehicle) as calculated in Eq.~\eqref{eq:final_reg_loss}. Since the Z-axis of bounding boxes is always set to be perpendicular to the horizontal plane, the distances of the closer surfaces between the predictions and ground truth are only related to the X and Y coordinates. We therefore remove the regression for $z_{\rm cv}$ and focus on $x_{\rm cv}$ and $y_{\rm cv}$. To avoid the overfitting problem on object sizes, we also remove the parts of regression loss for the residuals related to object sizes (\ie $l,w,h$). As a result, we only refine $x_{\rm cv}$, $y_{\rm cv}$, and $\theta$ in our EdgeHead and keep the $z_{\rm cv}$, $l,w$, and $h$ as predicted by the model's first stage. The new regression loss of our EdgeHead is therefore defined as
\begin{align}\label{eq:reg_loss_new}
    {\cal L}_{\rm reg}^{\prime} = \sum\limits_{r \in \left\{ x_{\rm cv}, y_{\rm cv}, \theta \right\} } {\cal L}_{{\rm smooth}-\ell_1}(\widehat{\Delta r^{a}}, \Delta r^{a}).
\end{align}

\begin{figure}[t]
\centering
   \includegraphics[width=0.95\linewidth]{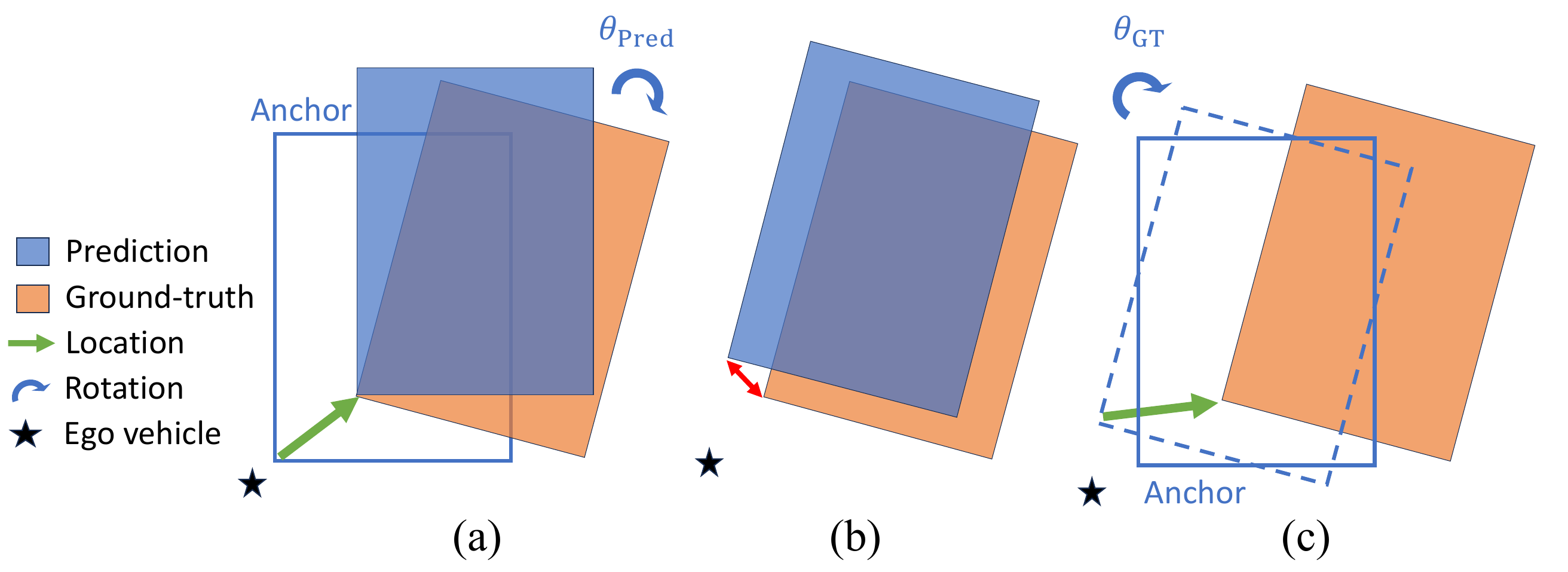}
\vspace{-0.08in}
\caption{\textbf{Illustration of different regression processes.} (a) The process that directly regresses the closest vertex and rotations without rotating the anchor box first. (b) The prediction obtained using the process in (a), in which the red arrow shows that the prediction does not learn the closest vertex as expected. (c) The regression process guided by Eq.~\eqref{eq:final_reg_loss} and Eq.~\eqref{eq:reg_loss_new} in our EdgeHead, which first rotates the anchor by $\theta_{\rm gt}$ and then calculate the regression target of $x$ and $y$ locations.}
\label{fig:long}
\label{fig:regress_corner}
\end{figure}


\noindent \textit{Remark.}
The rotation of the anchor box used in Eq.~\eqref{eq:final_reg_loss} is important in the modification of the regression process to realize our real purpose, \ie to guide the model to learn the closer-surfaces offsets between the predictions and ground truth. Considering an example of the model's regression process without the rotation as shown in Figure~\ref{fig:regress_corner}(a), the regression target related to $x$ and $y$ will guide the model to predict the residual so that the predicted box
can coincide with the ground-truth box at the vertex closest to the origin. However, since we are also regressing the rotation angle $\theta$, we will finally get a predicted box as shown in Figure~\ref{fig:regress_corner}(b), whose closest vertex to the origin does not coincide with the ground truth's anymore (see the red arrow). To consider the rotation regression as well, we first rotate the anchor box by the rotation angle $\theta_{\rm gt}$ of the ground-truth box as shown in Figure~\ref{fig:regress_corner}(c), and then calculate the residuals of $x_{\rm cv}$ and $y_{\rm cv}$ between the rotated anchor box and ground truth as the new regression target. Such a modified regression process makes the prediction box's closest vertex finally coincide with the ground-truth box's.

\subsection{Use of point-wise features in EdgeHead}\label{sec:pointfeaturehead}

We notice that some models use the additional raw point features as part of the input of the RoI refinement head to aggregate structural and spatial information into the RoI features to help improve object localization accuracy. This inspires us to investigate whether the point-wise features can also help improve the detection quality of the closer surfaces. To do so, we further include the point feature aggregation into the RoI pooling module of our EdgeHead. Specifically, we follow the idea of PV-RCNN~\cite{Shi_2020_CVPR} to extract the point-wise features. The keypoints are sampled from the original point cloud by the furthest-point-sampling algorithm, and the predicted-keypoint-weighting module is used to re-assign the weights of each point feature, which consists of a three-layer MLP network and a sigmoid function to predict the confidence that each point belongs to the foreground (\ie inside an object box). Afterwards, the weighted keypoint features are aggregated into the related RoIs via the set-abstraction-based RoI grid pooling, together with the 3D convolution features described in Section~\ref{sec:edgehead} above. We name this extended EdgeHead the \textbf{\textit{point-enhanced EdgeHead}} -- shortened as \textbf{\textit{PointEdgeHead}}.


\section{Experiments}
\label{sec:exp}

\begin{figure*}[h]
\centering
\includegraphics[width=0.95\textwidth]{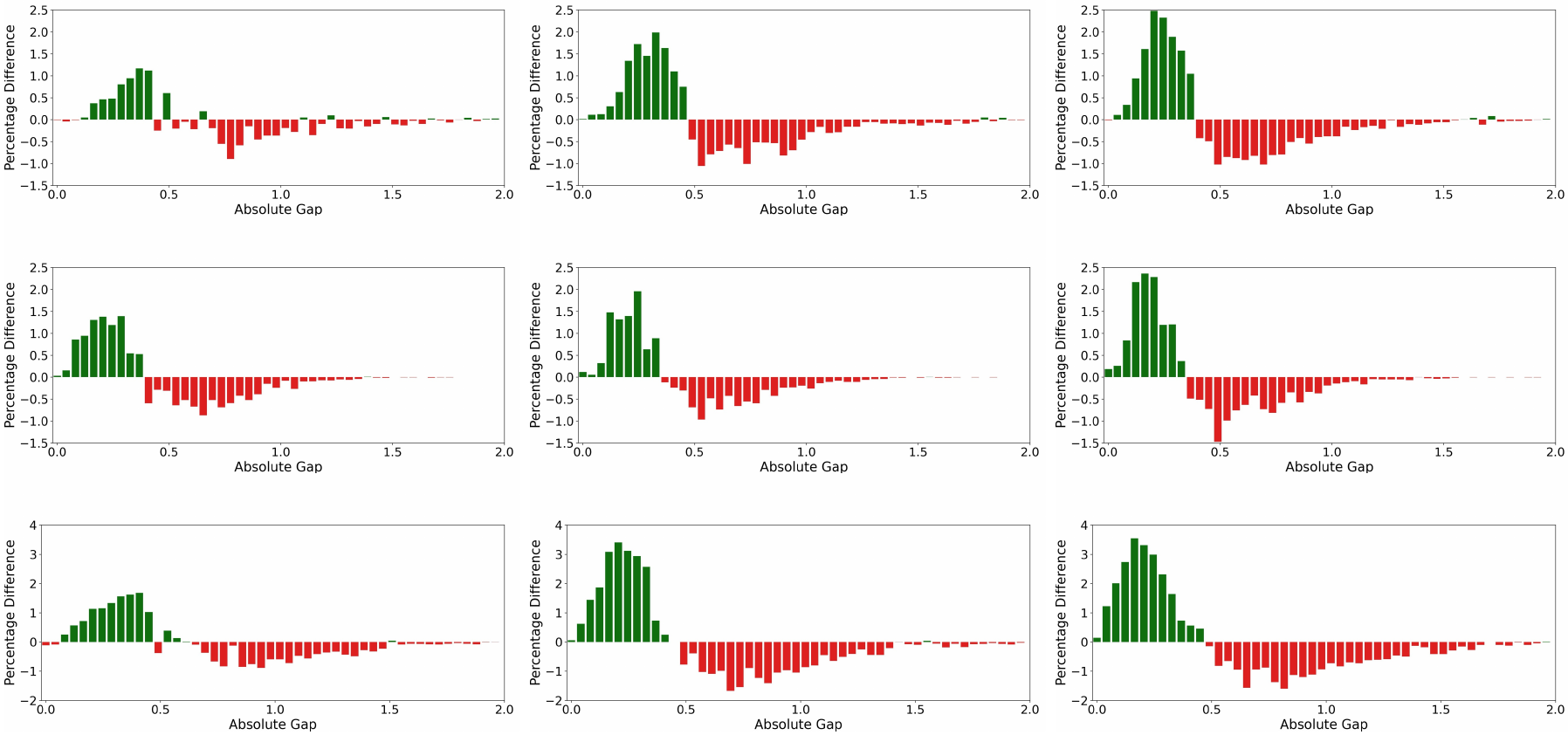}
\put(-460,152){\tiny (a) Waymo-KITTI: SECOND vs SECOND+EdgeHead}
\put(-305,152){\tiny (b) Waymo-KITTI: SECOND+ROS vs SECOND+ROS+EdgeHead}
\put(-140,152){\tiny (c) Waymo-KITTI: SECOND+SN vs SECOND+SN+EdgeHead}
\put(-460,72){\tiny (d) Waymo-nuScenes: CenterPoint vs CenterPoint+EdgeHead}
\put(-312,72){\tiny (e) Waymo-nuScenes: CenterPoint+ROS vs CenterPoint+ROS+EdgeHead}
\put(-147,72){\tiny (f) Waymo-nuScenes: CenterPoint+SN vs CenterPoint+SN+EdgeHead}
\put(-460,-8){\tiny (g) nuScenes-KITTI: SECOND vs SECOND+PointEdgeHead}
\put(-314,-8){\tiny (h) nuScenes-KITTI: SECOND+ROS vs SECOND+ROS+PointEdgeHead}
\put(-148,-8){\tiny (i) nuScenes-KITTI: SECOND+SN vs SECOND+SN+PointEdgeHead}
\caption{\textbf{Proportion difference of the absolute gap of the closer surfaces.} (a)--(c): SECOND in the Waymo $\rightarrow$ KITTI task. (d)--(f): CenterPoint in the Waymo $\rightarrow$ nuScenes task. (g)--(i): SECOND with our PointEdgeHead in the nuScenes $\rightarrow$ KITTI task. Columns one to three show the comparisons of models without domain adaptation methods, with the ROS, and with the SN, respectively. More results are given in the Supplementary Material.}
\label{fig:Gcs}
\end{figure*}
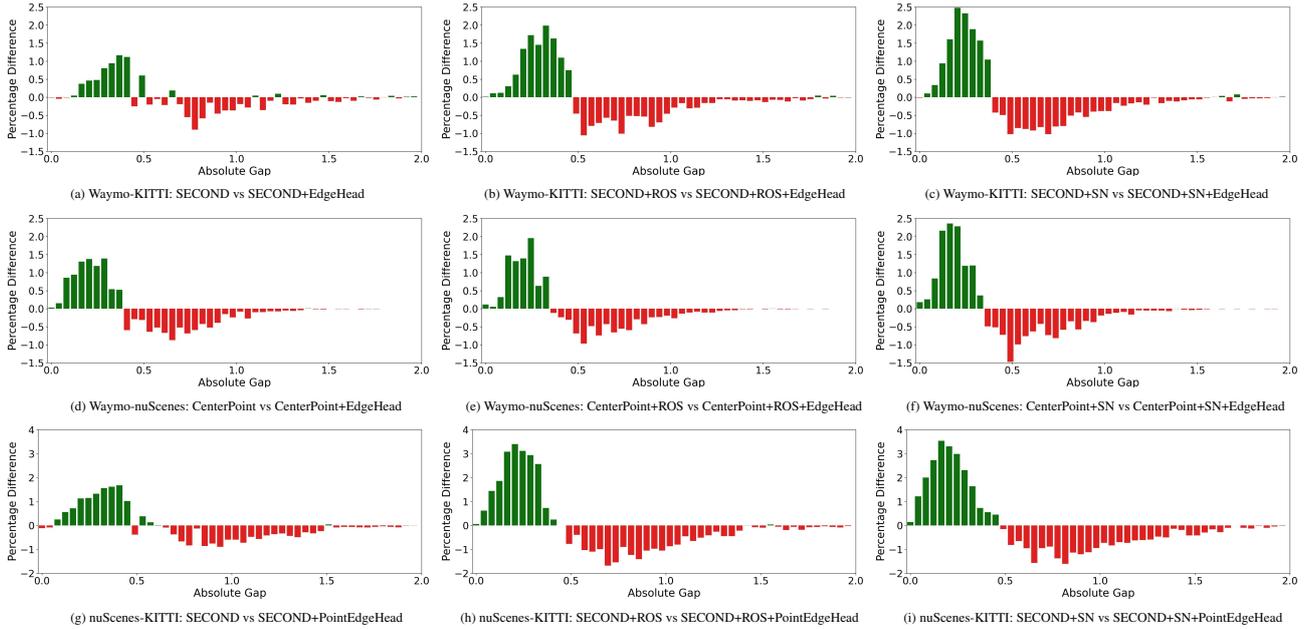

\subsection{Datasets and models}
\noindent\textbf{Datasets.}
We conduct our main experiments on three datasets that have been widely used in 3D object detection tasks: KITTI~\cite{6248074}, nuScenes~\cite{9156412}, and Waymo~\cite{Sun_2020_CVPR}. Following existing works~\cite{9156543, 9578132, wei2022lidar}, we use the KITTI evaluation metric for all datasets on the car category (\ie the vehicle category in Waymo). As mentioned in~\cite{9578132}, KITTI only provides annotations in the front view, which makes it much more difficult to adapt models from KITTI to the other two datasets that provide ring view point cloud data and annotations. We therefore evaluate the models' cross-domain performance via the following tasks: nuScenes $\rightarrow$ KITTI, Waymo $\rightarrow$ KITTI and Waymo $\rightarrow$ nuScenes.



\noindent\textbf{Data integration.} 
Most existing 3D object detection methods~\cite{s18103337, Shi_2020_CVPR, 8954080} aimed to achieve higher performance within each specific domain/dataset, and they often fine-tuned the models for different datasets independently (\eg adjusting the hyper-parameters related to data preprocessing and using different voxel sizes) without considering the influence of domain gaps. However, to investigate the cross-domain performance of these models, we must find a way to merge these datasets. We note that the following differences between datasets have a significant influence on the cross-domain experiments, \ie (i) the point cloud range; (ii) the origin of coordinates; and (iii) the unit for preprocessing the point cloud data, such as voxel sizes in voxel-based methods. Following the ideas in previous works~\cite{9156543, 9578132}, some preprocessing methods are adopted. We set the point cloud range of all datasets to $[-75.2, -75.2, -2, 75.2, 75.2, 4]$ meters and shift the whole point cloud space of different datasets vertically so that the X-Y plane always coincides with the horizontal plane. Following~\cite{9578132}, we set the voxel size of voxel-based methods to $(0.1, 0.1, 0.15)$ meters for all datasets.

\noindent\textbf{Baseline models.}
To better investigate the influence of model structures on cross-domain performance, we train SECOND~\cite{s18103337} and CenterPoint~\cite{Yin_2021_CVPR} on KITTI, Waymo, and nuScenes using the OpenPCDet~\cite{openpcdet2020} toolbox with suggested numbers of epochs and learning rates. In detail, we train them on KITTI for 80 epochs with learning rate $1\times10^{-3}$ and batch size of 8. Epochs of 50 and batch size of 16 are used for the training of the same models on nuScenes, and epochs of 30 and batch size of 8 for Waymo. For the models equipped with our EdgeHead, we train them based on the pre-trained original models for the same epochs as above, during which the parameters of the original models are frozen and only the heads are being trained. We also train the above models with two domain adaptation methods -- ROS~\cite{9578132} and SN~\cite{9156543} -- and combine them with our EdgeHead. We use the same training settings as their original reproductions on OpenPCDet.  Following other works~\cite{ bai2021pointdsc, 9578132} based on OpenPCDet, we adopt random horizontal flip, rotation, and scale transforms during the training process. All  models are trained on RTX 8000.

\begin{table}[t]
\caption{\textbf{Performance comparison of original models under four metrics in both cross-domain and within-domain tasks.} W, K, and N represent the Waymo, KITTI, and nuScenes datasets, respectively.}
\label{tab:Original models}
\centering
\begin{tabular}{c c c c c c} 
\toprule
Task & Method & \APBEV & \APthreeD & \APnew & \APcs \\
\midrule
\multirow{2}{*}{W $\rightarrow$ K}  
& SECOND & 49.2 & 9.3 & 19.0 & 10.9 \\
& CenterPoint & 51.3 & 13.1 & 18.2 & 9.5 \\
\midrule
\multirow{2}{*}{W $\rightarrow$ N}    
& SECOND & 27.8 & 16.1 & 15.6 & 6.7 \\
& CenterPoint & 30.4 & 16.7 & 19.9 & 12.3 \\
\midrule
\multirow{2}{*}{N $\rightarrow$ K} 
& SECOND & 35.7 & 11.8 & 16.4 & 9.8 \\
& CenterPoint & 34.6 & 8.3 & 13.1 & 5.8 \\
\midrule
\multirow{2}{*}{K $\rightarrow$ K} 
& SECOND & 84.3 & 72.1 & 71.4 & 53.3 \\
& CenterPoint & 84.4 & 73.6 & 72.8 & 56.8 \\
\bottomrule
\end{tabular}
\end{table}

\noindent\textbf{Evaluation metrics.}
We follow KITTI to evaluate the models' performance under the original BEV and 3D metrics. The evaluation is focused on the \textit{Car} category (\ie the \textit{Vehicle} in Waymo) which has the most samples in all the datasets and has been the main focus in existing works. AP (average precision) for BEV and 3D metrics (\ie the \APBEV and \APthreeD) with the IoU threshold at 0.7 is reported, \ie a car is marked as correctly detected if the IoU between the prediction and the ground truth is larger than 0.7. Our proposed metrics, \ie the CS-ABS AP and the CS-BEV AP, are denoted by  \APcs and \APnew, respectively; and we report the results with the IoU threshold at 0.7 for \APcs and 0.5 for \APnew (given that \APnew is more difficult to reach 0.7). 

\begin{table*}[htbp]
    \caption{\textbf{Main comparisons for SECOND and CenterPoint across different tasks.} We report \APBEV,  \APthreeD and \APcs of the car category at IoU = 0.7 and \APnew at IoU = 0.5. The reported performance is the moderate case when KITTI is the target domain, and is the overall result for other cross-domain tasks. Improvement (\ie fifth column) is calculated by the relative difference between each used method and the original model (\ie the first row of each task). }
    \label{tab:main_comparison}
    \setlength\tabcolsep{5pt}
    \centering
        \begin{tabular}{c c c c c c c c}
            \toprule[1pt]
            \multirow{3}{*}{Task} & \multirow{3}{*}{Method}  & \multicolumn{3}{c}{SECOND} & \multicolumn{3}{c}{CenterPoint} \\
            &  & \APBEV / \APthreeD & \APnew / \APcs & Improvement (\%) & \APBEV / \APthreeD & \APnew / \APcs & Improvement (\%) \\
            \midrule
            \multirow{6}{*}{W $\rightarrow$ K} 
            & Original  & 49.2 / 9.3 & 19.0 / 10.9 & - & 51.3 / 13.1 & 18.2 / 9.5 & - \\
            & + EdgeHead & 52.3 / 10.7 & 23.7 / 14.7 & 24.7\% / 34.9\% & 53.9 / 14.5 & 22.0 / 13.3 & 20.9\% / 40.0\% \\
            & + ROS & 73.0 / 38.3 & 33.7 / 12.6 & 77.4\% / 15.6\% & 75.1 / 44.2 & 41.1 / 19.1 & 126.1\% / 101.1\% \\
            & + EdgeHead \& ROS & 76.4 / 41.5 & 42.9 / 20.4 & 125.8\% / 87.2\% & 77.3 / 47.4 & 46.2 / 23.2 & 154.4\% / 144.2\% \\
            & + SN & 73.0 / 55.5 & 49.3 / 20.5 & 159.5\% / 87.9\% & 72.5 / 56.7 & 51.4 / 24.9 & 182.4\% / 162.1\% \\
            & + EdgeHead \& SN & 79.7 / 64.2 & 62.3 / 34.2 & 227.9\% / 213.8\% & 77.8 / 63.5 & 59.8 / 30.1 & 228.6\% / 216.8\% \\
            \midrule
            \multirow{6}{*}{W $\rightarrow$ N} 
            & Original  & 27.8 / 16.1 & 15.6 / 6.7 & - & 30.4 / 16.7 & 19.9 / 12.3 & - \\
            & + EdgeHead & 29.9 / 18.0 & 20.9 / 13.0 & 34.0\% / 94.0\% & 29.7 / 17.6 & 21.3 / 13.8 & 7.0\% / 12.2\% \\
            & + ROS & 26.7 / 15.4 & 15.8 / 6.5 & 1.3\% / -3.0\% & 28.8 / 16.2 & 19.5 / 11.7 & -2.0\% / -4.9\% \\
            & + EdgeHead \& ROS & 28.3 / 17.1 & 19.9 / 11.9 & 27.6\% / 77.6\% & 29.2 / 17.4 & 21.3 / 13.4 & 7.0\% / 9.3\% \\
            & + SN & 26.4 / 16.4 & 16.7 / 8.7 & 7.1\% / 29.9\% & 29.4 / 18.0 & 20.5 / 12.7 & 3.0\% / 3.3\% \\
            & + EdgeHead \& SN & 28.4 / 18.6 & 20.7 / 13.4 & 32.7\% / 100.0\% & 29.3 / 19.2 & 22.1 / 18.9 & 11.1\% / 53.7\% \\
            \midrule
            \multirow{6}{*}{N $\rightarrow$ K} 
            & Original  & 35.7 / 11.8 & 16.4 / 9.8 & - & 34.6 / 8.3 & 13.1 / 5.8 & - \\
            & + EdgeHead & 53.6 / 15.9 & 33.3 / 19.6 & 103.0\% / 100.0\% & 37.0 / 10.4 & 19.6 / 11.5 & 49.6\% / 98.3\% \\
            & + ROS & 43.4 / 20.0 &  20.2 / 8.1 & 23.2\% / -17.3\% & 43.8 / 20.6 & 27.6 / 13.1 & 110.8\% / 125.9\% \\
            & + EdgeHead \& ROS & 52.7 / 33.1 & 39.9 / 24.6 & 143.3\% / 151.0\% & 60.3 / 31.3 & 43.2 / 21.3 & 229.8\% / 267.2\% \\
            & + SN & 29.6 / 14.3 & 15.7 / 8.2 & -4.3\% / -16.3\% & 33.5 / 18.1 & 22.0 / 11.6 & 67.9\% / 100.0\% \\
            & + EdgeHead \& SN & 45.7 / 30.4 & 35.1 / 23.5 & 114.0\% / 139.8\% & 58.4 / 34.7 & 44.8 / 26.8 & 241.2\% / 362.1\% \\
            \bottomrule[0.8pt]
        \end{tabular}
\end{table*}

\subsection{The absolute gap of the closer surfaces}

We first compare the proposed EdgeHead with existing methods by measuring their absolute gaps of the closer surfaces (\ie $G_{\rm cs}$). As shown in Figure~\ref{fig:Gcs}, we calculate the distributions of $G_{cs}$ for each comparison pair of methods and draw the proportion difference between them. Specifically, we quantify the $G_{\rm cs}$ distribution of two models within an identical interval $I$ (set to $[0,2]$ by default), and then calculate the proportion difference as
$
    {\rm Diff}_{AB}(i) = P_B^i - P_A^i,
$
where $P_{A}^i$ and $P_{B}^i$ denote the proportion of $G_{\rm cs}$ in the $i\text{-th}$ sub-interval of $I$ for models $A$ and $B$, respectively. Therefore, if the left part of the proportion difference graph is above the X-axis and the right half is vice versa, we can tell that model B predicts the closer surfaces better than model A and thus has a $G_{\rm cs}$ distribution closer to zero. For example, Figure~\ref{fig:Gcs}(a) shows that SECOND+EdgeHead (\ie the SECOND model combined with our proposed EdgeHead) predicts the closer surfaces better than the original SECOND model when trained on Waymo and tested on KITTI. Consistent results are observed for the other tasks and models, which demonstrates that our EdgeHead can stably shift the $G_{cs}$ distribution to the left, \ie improve the detection quality regarding the closer surfaces. 

We also plot the proportion difference to analyze models using domain adaptation methods (\ie ROS and SN; see the last two columns in Figure~\ref{fig:Gcs}), and using our PointEdgeHead (see the last row in Figure~\ref{fig:Gcs}), which will be further discussed in Sections~\ref{sec:rossn} and \ref{sec:pointenhanced_results}.


\subsection{Main results of the proposed EdgeHead}\label{sec:maincompare}

In this section, we extensively evaluate the models' performance before and after equipping with our EdgeHead and under different types of metrics including our proposed CS-ABS and CS-BEV metrics. We first compare the results of different models under the original BEV and 3D metrics with our proposed CS-ABS and CS-BEV metrics in Table~\ref{tab:Original models} to analyze the robustness of our new metrics. Then we analyze the performance of models before and after equipping with our proposed EdgeHead, and investigate the influence of using  EdgeHead and two domain adaptation methods simultaneously in Table~\ref{tab:main_comparison}. We also calculate the \textit{\textbf{Improvement}} value as the relative difference between the used methods (\eg + EdgeHead \& ROS) and the original models under the CS-ABS and CS-BEV metrics. Due to page limit, additional results are attached in the {\it Supplementary Material}.

\subsubsection{Evaluation on the original models}


First of all, we compare the performance of two existing models, \ie SECOND and CenterPoint under four metrics as shown in Table~\ref{tab:Original models}. We evaluate them on three cross-domain tasks and one within-domain task on KITTI. 
Table~\ref{tab:Original models} shows that our proposed CS-ABS and CS-BEV metrics provide results of a similar quantity level with the original BEV and 3D metrics for various models and tasks. Our metrics also show different characteristics compared with the original metrics. Taking the W $\rightarrow$ K (\ie Waymo $\rightarrow$ KITTI) task as an example, the BEV AP and 3D AP of SECOND are both lower than CenterPoint, but both the CS-ABS AP and CS-BEV AP of SECOND are higher. Therefore, when trained on Waymo and tested on KITTI, the SECOND model predicts the closer surfaces better than CenterPoint. This advantage, however, is obscured by its lower BEV AP and 3D AP scores before. Such results highlight the distinction  and importance of our closer-surfaces-based evaluation metrics against the traditional ones.

\subsubsection{Improvement by our EdgeHead}

Table~\ref{tab:main_comparison} presents the quantitative comparison of the performance between difffernt models before and after equipping with our proposed EdgeHead. It shows that models equipped with EdgeHead can achieve better CS-ABS AP and CS-BEV AP than the original models in all cross-domain tasks. As described in Section~\ref{sec:metricdesign}, the CS-ABS AP directly measures the improvement in the detection quality of the closer surfaces, and the consistently improved performance under this metric shows that EdgeHead can stably improve the closer-surfaces detection ability of existing models across various domains. {The results of the CS-ABS AP are also consistent with the proportion difference of the closer-surfaces absolute gap shown in Figure~\ref{fig:Gcs}.} Meanwhile, the improvement under the CS-BEV metric shows that EdgeHead also works well when evaluating models with a balance between the accuracy of the entire box and the closer surfaces.


Table~\ref{tab:main_comparison} also shows that the models' performance changes much less or even remains at the original value level when evaluated under the original BEV and 3D metrics. Taking the SECOND model in the Waymo $\rightarrow$ KITTI task as an example, the BEV AP and 3D AP respectively improved by 6.3\% (from 49.2 to 52.3) and 15.1\% (from 9.3 to 10.7) when equipped with EdgeHead, while the CS-ABS AP and CS-BEV AP represent the improvement by 24.7\% and 34.9\%, respectively. We also summarize the gaps between each original model before and after equipping with EdgeHead in Table~\ref{tab:main_comparison}, which shows that similar phenomena can also be observed in other comparisons. The larger improvement shown in the CS-ABS AP and CS-BEV AP supports two important conclusions: (i) the newly proposed metrics are truly more sensitive to the closer-surfaces detection ability, and therefore can evaluate the model's cross-domain performance from a different point of view; and (ii) our EdgeHead can effectively improve the model's ability to detect the closer surfaces, which is truly helpful for applications in cross-domain tasks.

\subsubsection{Combination with ROS and SN}\label{sec:rossn}

ROS~\cite{9578132} and SN~\cite{9156543} are two domain adaptation methods that aim to solve the overfitting problems in object sizes. ROS randomly scales the size of object boxes in both the annotations and the point cloud data to make the model more robust to object sizes. SN uses the average object size of each dataset as additional information and normalizes the source domain's object size by using the target domain's size statistics. It is therefore worth investigating the influence of such methods on the models' closer-surfaces detection ability. 

We below first evaluate the CS-ABS AP and CS-BEV AP of different models equipped with these two methods (\ie ROS and SN) including further combining them with our EdgeHead. We denote these combinations as +ROS, +SN, +EdgeHead \& ROS and +EdgeHead \& SN in Table~\ref{tab:main_comparison}. The comparisons of the absolute gap are shown in Figure~\ref{fig:Gcs} as well. For most tasks, ROS and SN can help the models achieve higher BEV AP and 3D AP, but cannot stably improve the CS-ABS AP and CS-BEV AP by a similar margin. Taking SECOND in the Waymo $\rightarrow$ KITTI task as an example, ROS increases the BEV AP and 3D AP respectively by 48.4\% and 311.8\% (\ie from 49.2 / 9.3 to 73.0 / 38.3) but only increases the CS-ABS AP by 15.6\%. In comparison, the additional use of our EdgeHead increases the performance under all four metrics, especially for the CS-ABS AP and CS-BEV AP. Taking the above example, the performance under the new metrics increases by 125.8\% and 87.2\% when equipping SECOND with ROS and EdgeHead simultaneously. Consistent results can also be observed for the other tasks and models in Table~\ref{tab:main_comparison} and Figure~\ref{fig:Gcs}.
%
We also noticed that for both models in the Waymo $\rightarrow$ nuScenes task, ROS and SN increase the performance much less or even decrease it due to the minor object size difference between these two datasets, which is also mentioned in \cite{9578132}. However, the performance can still be greatly improved by using our EdgeHead and ROS / SN together.

The above results demonstrate that our proposed EdgeHead can be effectively used with the existing domain adaptation methods designed for the size overfitting problem, which not only further improves the models' detection ability for the entire box but also helps achieve much better closer-surfaces detection quality compared with only using the existing domain adaptation methods.

\subsubsection{Influence of additional point-wise features}\label{sec:pointenhanced_results}

We now compare the performance between our proposed EdgeHead and its extended version PointEdgeHead taking additional point-wise features as described in Section~\ref{sec:pointfeaturehead}, see the results in Table~\ref{tab:pointedgehead} and the third column of Figure~\ref{fig:Gcs} where the SECOND model is utilized. Using additional point-wise features further improves the models' performance under the CS-ABS and CS-BEV metrics for most tasks, showing the point features' structural information can indeed be helpful for  the closer surfaces detection. The improvement is more obvious in the nuScenes $\rightarrow$ KITTI task, which indicates that the point-wise features may play an important role when adapting models from a sparser domain to a denser domain. The improvement is however not that obvious for the Waymo $\rightarrow$ KITTI and Waymo $\rightarrow$ nuScenes tasks, and sometimes using PointEdgeHead could lead to lower performance (\eg Waymo $\rightarrow$ nuScenes without ROS or SN). Considering the extra time and resources consumed by the point feature aggregation process, it is thus unnecessary to always consider PointEdgeHead. In particular, these results show that our EdgeHead has already effectively enhanced the models' closer-surfaces detection ability with high utilization of current input information.

\begin{table}[tbp]
\caption{\textbf{Comparison between our EdgeHead and PointEdgeHead under the SECOND model.} }
\label{tab:pointedgehead}
\centering
\begin{tabular}{c c c c c c } 
\toprule
Task                         & Method & + EdgeHead & + PointEdgeHead \\ 
            &   & \APnew / \APcs  & \APnew / \APcs \\
\midrule
\multirow{3}{*}{W $\rightarrow$ K}  
     & Original  &  23.7 / 14.7   & 24.4 / 16.9 \\
     & + ROS     & 42.9 / 20.4  & 47.2 / 25.0  \\
     & + SN    &  62.3 / 34.2  & 59.0 / 33.3  \\ 
     
\midrule
\multirow{3}{*}{W $\rightarrow$ N}    
    & Original  &     20.9 / 13.0   &  20.8 / 12.2  \\
    & + ROS  &     19.9 / 11.9     &  21.0 / 12.7  \\
    & + SN   & 20.7 / 13.4   &  22.0 / 14.3 \\ 
    
\midrule
\multirow{3}{*}{N $\rightarrow$ K} 
    & Original  &  33.3 / 19.6 &   36.5 / 21.0    \\
    & + ROS  &  39.9 / 24.6 &  46.2 / 27.3      \\
    & + SN &   35.1 / 23.5  &   43.6 / 28.3  \\

\bottomrule
\end{tabular}
\centering
\end{table}

\subsection{Ablation study}


\begin{table}[tbp]
\caption{\textbf{Ablation study of our EdgeHead under the SECOND model.} Refine: Using a second stage refinement head. Corner: Replacing the center locations with the locations of the closest vertex in the refinement head. }
\setlength\tabcolsep{4.5pt}
\label{tab:controlgroup}
\centering
\begin{tabular}{c c c c c c} 
\toprule
Method                         & Refine & Corner & W $\rightarrow$ K & W $\rightarrow$ N & N $\rightarrow$ K\\ 
\midrule
     Original  &  \XSolidBrush & \XSolidBrush & 19.0 / 10.9   & 15.6 / 6.7 & 16.4 / 9.8 \\

     Control-group  & \Checkmark & \XSolidBrush & 21.6 / 12.1   & 18.7 / 11.4 & 19.2 / 10.1 \\

     EdgeHead  & \Checkmark & \Checkmark &  23.7 / 14.7   & 20.9 / 13.0 & 33.3 / 19.6 \\
     
\bottomrule
\end{tabular}
\centering
\end{table}

To further analyze our EdgeHead's refinement performance, below we propose another control-group head for us to conduct ablation study. Specifically, we maintain the module structure and the loss function design of EdgeHead, while replacing the closest vertex in EdgeHead with the center coordinates for the calculation of the location regression target. Therefore, the loss function of the control-group head reads
\begin{align}
    {\cal L}_{\rm reg}^{\prime \prime} = \sum\limits_{r \in \left\{ x_{\rm c}, y_{\rm c}, \theta \right\} } {\cal L}_{{\rm smooth}-\ell_1}(\widehat{\Delta r^{a}}, \Delta r^{a}).
\end{align}
In other words, the above control-group head is a simplified version of the typical refinement module as described in Eq.~\eqref{eq:origin_loss}, which only refines the (BEV) location $x$, $y$, and the rotation angle $\theta$. By comparing the performance of EdgeHead and this control-group head, we can better understand the contribution of modifying the training purpose to the closer surfaces. As shown in Table~\ref{tab:controlgroup}, although the performance of the control-group head is better than the original model that does not use any refinement head, there is a rather significant gap in terms of detection performance improvement when comparing to the excellent results of our EdgeHead. The results in Table~\ref{tab:controlgroup} indicate that the regression target in our EdgeHead truly helps models achieve better closer-surfaces detection ability.

\section{Conclusion}

In this paper, we novelly view the cross-domain 3D object detection problem from the detection quality of the closer surfaces to the ego vehicle. We proposed two evaluation metrics, \ie the CS-ABS AP and CS-BEV AP, to measure this detection quality and achieve a balance between the entire boxes and the closer surfaces of the objects. The proposed metrics are less sensitive to the object size difference among datasets and thus can evaluate the models' performance across domains more reasonably. Meanwhile, we equipped the existing models with our proposed EdgeHead to guide them to focus more on the closer-surfaces gaps during training. Extensive experiments show that EdgeHead can effectively help models  detect better the closer surfaces and perform better under both the existing metrics and our proposed metrics. The results indicate that by guiding models to focus more on the surfaces with more points captured by the LiDAR sensor, the models can learn more robust knowledge from the training domain and perform better in cross-domain tasks.

\begin{ack}
We express sincere gratitude to Xiangyu Chen and Runwei Guan for their inspiration at the early stage of this project.
\end{ack}



\bibliography{ecai-sample-and-instructions}

\begin{thebibliography}{10}

\bibitem{Bai_2022_CVPR}
Xuyang Bai, Zeyu Hu, Xinge Zhu, Qingqiu Huang, Yilun Chen, Hongbo Fu, and Chiew-Lan Tai, `Transfusion: Robust lidar-camera fusion for 3d object detection with transformers', in {\em Proceedings of the IEEE/CVF Conference on Computer Vision and Pattern Recognition (CVPR)}, pp. 1090--1099, (June 2022).

\bibitem{9156412}
Holger Caesar, Varun Bankiti, Alex~H. Lang, Sourabh Vora, Venice~Erin Liong, Qiang Xu, Anush Krishnan, Yu~Pan, Giancarlo Baldan, and Oscar Beijbom, `nuscenes: A multimodal dataset for autonomous driving', in {\em 2020 IEEE/CVF Conference on Computer Vision and Pattern Recognition (CVPR)}, pp. 11618--11628, (2020).

\bibitem{carion2020end}
Nicolas Carion, Francisco Massa, Gabriel Synnaeve, Nicolas Usunier, Alexander Kirillov, and Sergey Zagoruyko, `End-to-end object detection with transformers', in {\em European conference on computer vision}, pp. 213--229. Springer, (2020).

\bibitem{8100174}
Xiaozhi Chen, Huimin Ma, Ji~Wan, Bo~Li, and Tian Xia, `Multi-view 3d object detection network for autonomous driving', in {\em 2017 IEEE Conference on Computer Vision and Pattern Recognition (CVPR)}, pp. 6526--6534, (2017).

\bibitem{8578921}
Yuhua Chen, Wen Li, and Luc~Van Gool, `Road: Reality oriented adaptation for semantic segmentation of urban scenes', in {\em 2018 IEEE/CVF Conference on Computer Vision and Pattern Recognition}, pp. 7892--7901, (2018).

\bibitem{chen2023voxenext}
Yukang Chen, Jianhui Liu, Xiangyu Zhang, Xiaojuan Qi, and Jiaya Jia, `Voxelnext: Fully sparse voxelnet for 3d object detection and tracking', in {\em Proceedings of the IEEE/CVF Conference on Computer Vision and Pattern Recognition}, (2023).

\bibitem{chen2023revisiting}
Zhuoxiao Chen, Yadan Luo, Zheng Wang, Mahsa Baktashmotlagh, and Zi~Huang, `Revisiting domain-adaptive 3d object detection by reliable, diverse and class-balanced pseudo-labeling', in {\em Proceedings of the IEEE/CVF International Conference on Computer Vision}, pp. 3714--3726, (2023).

\bibitem{deng2020voxel}
Jiajun Deng, Shaoshuai Shi, Peiwei Li, Wengang Zhou, Yanyong Zhang, and Houqiang Li, `Voxel r-cnn: Towards high performance voxel-based 3d object detection', {\em arXiv:2012.15712}, (2020).

\bibitem{dosovitskiy2020vit}
Alexey Dosovitskiy, Lucas Beyer, Alexander Kolesnikov, Dirk Weissenborn, Xiaohua Zhai, Thomas Unterthiner, Mostafa Dehghani, Matthias Minderer, Georg Heigold, Sylvain Gelly, Jakob Uszkoreit, and Neil Houlsby, `An image is worth 16x16 words: Transformers for image recognition at scale', {\em ICLR}, (2021).

\bibitem{6248074}
Andreas Geiger, Philip Lenz, and Raquel Urtasun, `Are we ready for autonomous driving? the kitti vision benchmark suite', in {\em 2012 IEEE Conference on Computer Vision and Pattern Recognition}, pp. 3354--3361, (2012).

\bibitem{7410526}
Ross Girshick, `Fast r-cnn', in {\em 2015 IEEE International Conference on Computer Vision (ICCV)}, pp. 1440--1448, (2015).

\bibitem{pmlr-v80-hoffman18a}
Judy Hoffman, Eric Tzeng, Taesung Park, Jun-Yan Zhu, Phillip Isola, Kate Saenko, Alexei Efros, and Trevor Darrell, `{C}y{CADA}: Cycle-consistent adversarial domain adaptation', in {\em Proceedings of the 35th International Conference on Machine Learning}, eds., Jennifer Dy and Andreas Krause, volume~80 of {\em Proceedings of Machine Learning Research}, pp. 1989--1998. PMLR, (10--15 Jul 2018).

\bibitem{Hsu_2020_WACV}
Han-Kai Hsu, Chun-Han Yao, Yi-Hsuan Tsai, Wei-Chih Hung, Hung-Yu Tseng, Maneesh Singh, and Ming-Hsuan Yang, `Progressive domain adaptation for object detection', in {\em Proceedings of the IEEE/CVF Winter Conference on Applications of Computer Vision (WACV)}, (March 2020).

\bibitem{hu2023density}
Qianjiang Hu, Daizong Liu, and Wei Hu, `Density-insensitive unsupervised domain adaption on 3d object detection', in {\em Proceedings of the IEEE/CVF Conference on Computer Vision and Pattern Recognition}, (2023).

\bibitem{Huang_2018_ECCV}
Haoshuo Huang, Qixing Huang, and Philipp Krahenbuhl, `Domain transfer through deep activation matching', in {\em Proceedings of the European Conference on Computer Vision (ECCV)}, (September 2018).

\bibitem{9008383}
Mehran Khodabandeh, Arash Vahdat, Mani Ranjbar, and William Macready, `A robust learning approach to domain adaptive object detection', in {\em 2019 IEEE/CVF International Conference on Computer Vision (ICCV)}, pp. 480--490, (2019).

\bibitem{li2023pillarnext}
Jinyu Li, Chenxu Luo, and Xiaodong Yang, `Pillarnext: Rethinking network designs for 3d object detection in lidar point clouds', in {\em IEEE/CVF Conference on Computer Vision and Pattern Recognition (CVPR)}, (2023).

\bibitem{NIPS2017_d8bf84be}
Charles~Ruizhongtai Qi, Li~Yi, Hao Su, and Leonidas~J Guibas, `Pointnet++: Deep hierarchical feature learning on point sets in a metric space', in {\em Advances in Neural Information Processing Systems}, eds., I.~Guyon, U.~Von Luxburg, S.~Bengio, H.~Wallach, R.~Fergus, S.~Vishwanathan, and R.~Garnett, volume~30. Curran Associates, Inc., (2017).

\bibitem{Shi_2020_CVPR}
Shaoshuai Shi, Chaoxu Guo, Li~Jiang, Zhe Wang, Jianping Shi, Xiaogang Wang, and Hongsheng Li, `Pv-rcnn: Point-voxel feature set abstraction for 3d object detection', in {\em Proceedings of the IEEE/CVF Conference on Computer Vision and Pattern Recognition (CVPR)}, (June 2020).

\bibitem{8954080}
Shaoshuai Shi, Xiaogang Wang, and Hongsheng Li, `Pointrcnn: 3d object proposal generation and detection from point cloud', in {\em 2019 IEEE/CVF Conference on Computer Vision and Pattern Recognition (CVPR)}, pp. 770--779, (2019).

\bibitem{Sun_2020_CVPR}
Pei Sun, Henrik Kretzschmar, Xerxes Dotiwalla, Aurelien Chouard, Vijaysai Patnaik, Paul Tsui, James Guo, Yin Zhou, Yuning Chai, Benjamin Caine, Vijay Vasudevan, Wei Han, Jiquan Ngiam, Hang Zhao, Aleksei Timofeev, Scott Ettinger, Maxim Krivokon, Amy Gao, Aditya Joshi, Yu~Zhang, Jonathon Shlens, Zhifeng Chen, and Dragomir Anguelov, `Scalability in perception for autonomous driving: Waymo open dataset', in {\em Proceedings of the IEEE/CVF Conference on Computer Vision and Pattern Recognition (CVPR)}, (June 2020).

\bibitem{openpcdet2020}
OpenPCDet~Development Team.
\newblock Openpcdet: An open-source toolbox for 3d object detection from point clouds.
\newblock \url{https://github.com/open-mmlab/OpenPCDet}, 2020.

\bibitem{8953674}
Tao Wang, Xiaopeng Zhang, Li~Yuan, and Jiashi Feng, `Few-shot adaptive faster r-cnn', in {\em 2019 IEEE/CVF Conference on Computer Vision and Pattern Recognition (CVPR)}, pp. 7166--7175, (2019).

\bibitem{9156543}
Yan Wang, Xiangyu Chen, Yurong You, Li~Erran Li, Bharath Hariharan, Mark Campbell, Kilian~Q. Weinberger, and Wei-Lun Chao, `Train in germany, test in the usa: Making 3d object detectors generalize', in {\em 2020 IEEE/CVF Conference on Computer Vision and Pattern Recognition (CVPR)}, pp. 11710--11720, (2020).

\bibitem{wei2022lidar}
Yi~Wei, Zibu Wei, Yongming Rao, Jiaxin Li, Jie Zhou, and Jiwen Lu, `Lidar distillation: Bridging the beam-induced domain gap for 3d object detection', {\em arXiv preprint arXiv:2203.14956}, (2022).

\bibitem{wu2023towards}
Guile Wu, Tongtong Cao, Bingbing Liu, Xingxin Chen, and Yuan Ren, `Towards universal lidar-based 3d object detection by multi-domain knowledge transfer', in {\em Proceedings of the IEEE/CVF International Conference on Computer Vision}, pp. 8669--8678, (2023).

\bibitem{bai2021pointdsc}
Bai Xuyang, Hu~Zeyu, Zhu Xinge, Huang Qingqiu, Chen Yilun, Fu~Hongbo, and Chiew-Lan Tai, `{TransFusion}: {R}obust {L}idar-{C}amera {F}usion for {3}d {O}bject {D}etection with {T}ransformers', {\em CVPR}, (2022).

\bibitem{s18103337}
Yan Yan, Yuxing Mao, and Bo~Li, `Second: Sparsely embedded convolutional detection', {\em Sensors}, {\bf 18}(10), (2018).

\bibitem{8578896}
Bin Yang, Wenjie Luo, and Raquel Urtasun, `Pixor: Real-time 3d object detection from point clouds', in {\em 2018 IEEE/CVF Conference on Computer Vision and Pattern Recognition}, pp. 7652--7660, (2018).

\bibitem{9578132}
Jihan Yang, Shaoshuai Shi, Zhe Wang, Hongsheng Li, and Xiaojuan Qi, `St3d: Self-training for unsupervised domain adaptation on 3d object detection', in {\em 2021 IEEE/CVF Conference on Computer Vision and Pattern Recognition (CVPR)}, pp. 10363--10373, (2021).

\bibitem{yang2021st3d++}
Jihan Yang, Shaoshuai Shi, Zhe Wang, Hongsheng Li, and Xiaojuan Qi, `St3d++: Denoised self-training for unsupervised domain adaptation on 3d object detection', {\em IEEE Transactions on Pattern Analysis and Machine Intelligence}, (2022).

\bibitem{Yin_2021_CVPR}
Tianwei Yin, Xingyi Zhou, and Philipp Krahenbuhl, `Center-based 3d object detection and tracking', in {\em Proceedings of the IEEE/CVF Conference on Computer Vision and Pattern Recognition (CVPR)}, pp. 11784--11793, (June 2021).

\bibitem{yuan2023bi3d}
Jiakang Yuan, Bo~Zhang, Xiangchao Yan, Tao Chen, Botian Shi, Yikang Li, and Yu~Qiao, `Bi3d: Bi-domain active learning for cross-domain 3d object detection', in {\em Proceedings of the IEEE/CVF Conference on Computer Vision and Pattern Recognition}, pp. 15599--15608, (2023).

\bibitem{zhang2023uni3d}
Bo~Zhang, Jiakang Yuan, Botian Shi, Tao Chen, Yikang Li, and Yu~Qiao, `Uni3d: A unified baseline for multi-dataset 3d object detection', in {\em Proceedings of the IEEE/CVF Conference on Computer Vision and Pattern Recognition}, pp. 9253--9262, (2023).

\bibitem{8578570}
Yin Zhou and Oncel Tuzel, `Voxelnet: End-to-end learning for point cloud based 3d object detection', in {\em 2018 IEEE/CVF Conference on Computer Vision and Pattern Recognition}, pp. 4490--4499, (2018).

\end{thebibliography}

\clearpage

\appendix
\begin{center}
	\textbf{\Large Supplementary Material}
\end{center}

In this Supplementary Material, we provide below the additional details omitted in the main paper.

\begin{itemize}
    
    \item Section~\ref{appendixsec:moreGcsFig}: Further results regarding the proportion difference of the absolute gap of the closer surfaces.
    \item Section~\ref{appendixsec:edgehead}: Details and discussions on the design of our EdgeHead.
    \item  Section~\ref{appendixsec:withinresults}: Results of within-domain tasks under the new metrics.
    \item Section~\ref{appendixsec:visuals}: Visualizations of predictions by using models equipped with EdgeHead.
    \item Section~\ref{appendixsec:alpha}: Settings of $\alpha$ in the new metrics.
\end{itemize}

\section{Proportion difference of the absolute gap}\label{appendixsec:moreGcsFig}

We provide additional results of the proportion difference of the absolute gap of the closer surfaces, see the results of SECOND and CenterPoint respectively in Figure~\ref{fig:more_Gcs_second} and Figure~\ref{fig:more_Gcs_cp}, and see the  results of SECOND equipped with PointEdgeHead in Figure~\ref{fig:more_Gcs_point}.

\begin{figure*}[h]
\centering
\includegraphics[width=0.95\textwidth]{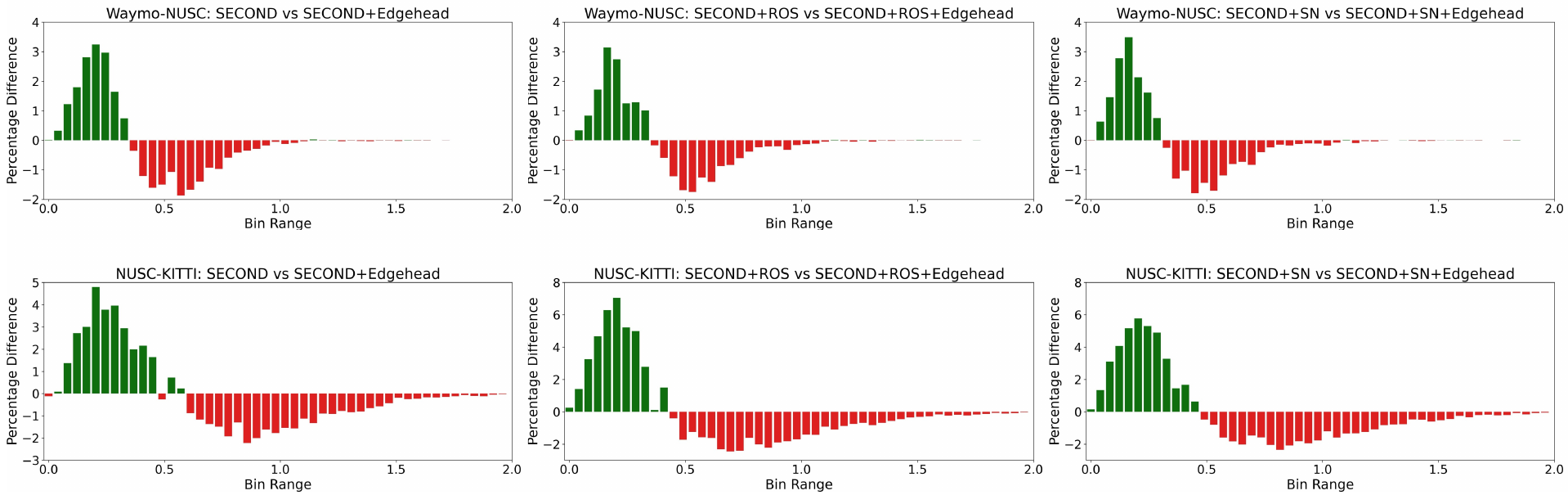}
\caption{\textbf{Proportion difference of the absolute gap of the closer surfaces.} Results for SECOND in Waymo $\rightarrow$ nuScenes and nuScenes $\rightarrow$ KITTI tasks.}
\label{fig:more_Gcs_second}
\end{figure*}

\begin{figure*}[h]
\centering
\includegraphics[width=0.95\textwidth]{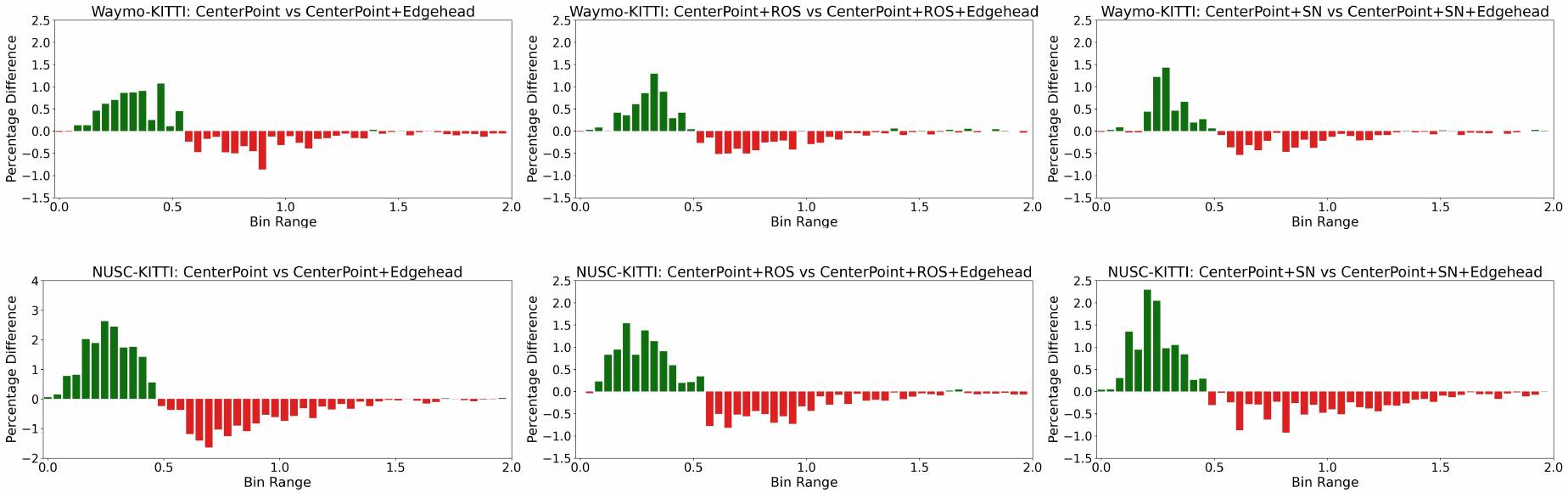}
\caption{\textbf{Proportion difference of the absolute gap of the closer surfaces.} Results for CenterPoint in Waymo $\rightarrow$ KITTI and nuScenes $\rightarrow$ KITTI tasks.}
\label{fig:more_Gcs_cp}
\end{figure*}

\begin{figure*}[h]
\centering
\includegraphics[width=0.95\textwidth]{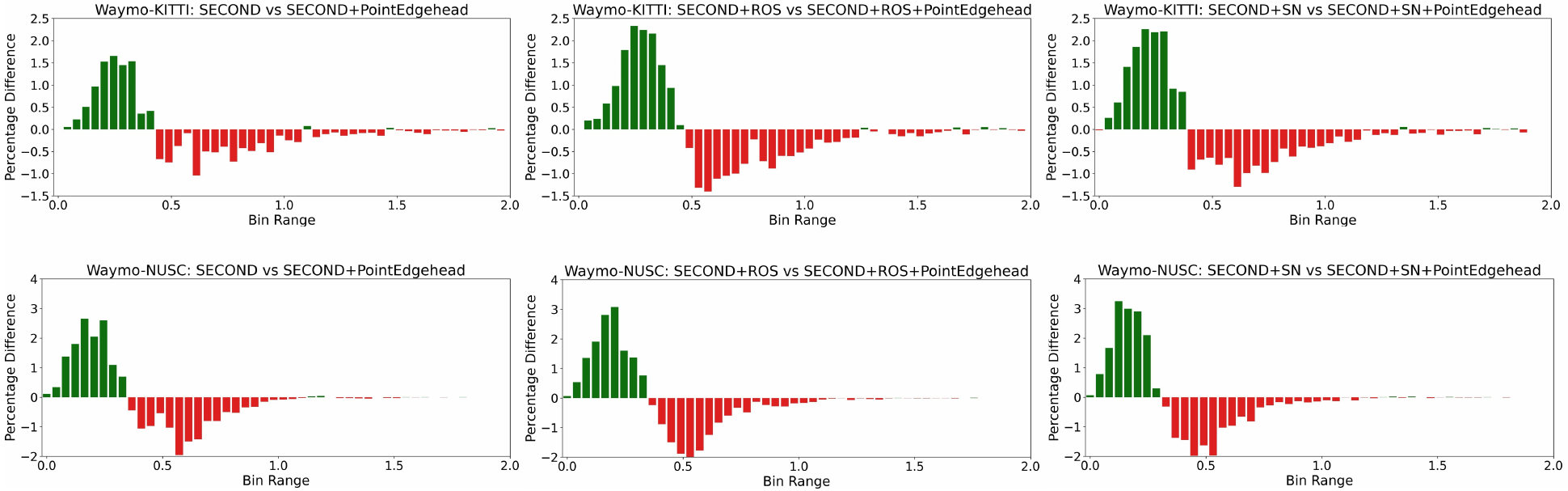}
\caption{\textbf{Proportion difference of the absolute gap of the closer surfaces.} Results for SECOND equipped with PointEdgeHead in Waymo $\rightarrow$ KITTI and Waymo $\rightarrow$ nuScenes tasks.}
\label{fig:more_Gcs_point}
\end{figure*}

\section{Box generation process in EdgeHead}\label{appendixsec:edgehead}
In the design of EdgeHead, we initially considered using ideas that generate the predictions from the box corners. However, we decide to keep the box generation process as the refinement stage of anchor-based methods due to the below two main reasons. 

Firstly, to generate boxes from the corner, we need to use the following information: (a) the coordinates of the corner; (b) the dimensions of the box, \ie the length, width, and height; and (c) the rotation angle of the box. This is similar with the required information when generating boxes based on anchors. However, the box generation from the corner requires one more input, \ie the relative position between the corner and the center of the box. As there are four corners for each box, four boxes can be generated for the same input as described above if we do not tell the relative position. This involves additional variables during the training process, making the structure of the refinement head more complex. Meanwhile, it provides more noise into the prediction of the boxes, since it is impossible that a model can 100\% correctly predict the relative position between the corner and the center. 
Secondly, the main purpose of our EdgeHead is to find an efficient and effective way to guide models to focus more on the detection quality of the closer surfaces instead of the center, we are therefore not intended to modify the structure of the existing models greatly. Using the corner-based box generation method, it is more natural to modify models from the first stage than to add an extra refinement head.

As a result, we decide to follow the anchor-based methods to generate boxes in our EdgeHead. It is also interesting to consider the corner-based box generation methods as a part of our future work.

\section{Results of within-domain tasks under the new metrics}\label{appendixsec:withinresults}

We provide additional results of the SECOND and CenterPoint models in within-domain tasks, including KITTI $\rightarrow$ KITTI, nuScenes $\rightarrow$ nuScenes, and Waymo $\rightarrow$ Waymo, see Tables \ref{tab:within-domain-second} and \ref{tab:within-domain-centerpoint}. Note that since ROS and SN are domain adaptation methods designed for cross-domain tasks, we skip evaluating the models equipped with them in within-domain tasks. The results show that equipping with our EdgeHead can also improve the models' performance in within-domain tasks, under both the original BEV or 3D metrics and the proposed CS-ABS and CS-BEV metrics.

\begin{table}[t]
\caption{\textbf{Performance of SECOND under four metrics in within-domain tasks.}}
\label{tab:within-domain-second}
\centering
\begin{tabular}{c c c c c c} 
\toprule
Task & Method & \APBEV / \APthreeD & \APnew / \APcs \\
\midrule
\multirow{2}{*}{K $\rightarrow$ K}  
& Original & 84.3 / 72.1 & 71.4 / 53.3 \\
& + EdgeHead & 88.2 / 77.5 & 79.6 / 66.7 \\
\midrule
\multirow{2}{*}{N $\rightarrow$ N}    
& Original & 43.1 / 22.6 & 25.9 / 14.2\\
& + EdgeHead & 46.7 / 29.7 & 34.6 / 25.2 \\
\midrule
\multirow{2}{*}{W $\rightarrow$ W} 
& Original & 63.1 / 47.5 & 47.3 / 34.2 \\
& + EdgeHead & 65.1 / 51.4 & 53.8 / 44.1 \\
\bottomrule
\end{tabular}
\end{table}

\begin{table}[t]
\caption{\textbf{Performance of CenterPoint under four metrics in within-domain tasks.}}
\label{tab:within-domain-centerpoint}
\centering
\begin{tabular}{c c c c c c} 
\toprule
Task & Method & \APBEV / \APthreeD & \APnew / \APcs \\
\midrule
\multirow{2}{*}{K $\rightarrow$ K}  
& Original & 84.4 / 73.6 & 72.8 / 56.8 \\
& + EdgeHead & 84.4 / 75.3 & 74.7 / 61.8 \\
\midrule
\multirow{2}{*}{N $\rightarrow$ N}    
& Original & 49.7 / 30.7 & 34.1 / 23.2 \\
& + EdgeHead & 48.1 / 32.5 & 36.9 / 26.9 \\
\midrule
\multirow{2}{*}{W $\rightarrow$ W} 
& Original & 67.3 / 52.5 & 52.6 / 41.4 \\
& + EdgeHead & 65.4 / 53.8 & 56.0 / 46.7 \\
\bottomrule
\end{tabular}
\end{table}

\begin{table}[htbp]
\caption{\textbf{Results of SECOND and SECOND equipped with EdgeHead under the CS-ABS and CS-BEV metrics with different $\alpha$ settings.}  The results on the Waymo $\rightarrow$ KITTI task are reported in this table. }

\label{tab:appendix_alpha}
\centering
\begin{tabular}{c c c c c c } 
\toprule
Tasks                         & Metrics & SECOND & SECOND + EdgeHead\\ 
            &   & \APnew / \APcs  & \APnew / \APcs \\
\midrule
\multirow{3}{*}{$\alpha = 0.5 $}  
     & Original  &  63.8 / 70.5   & 69.0 / 74.8 \\
     & + ROS     & 75.1 / 75.2  & 79.8 / 81.6  \\
     & + SN    &  72.8 / 72.2  & 80.0 / 80.0 \\ 
     
\midrule
\multirow{3}{*}{$\alpha = 1 $}    
    & Original  &     19.0 / 10.9   &  23.7 / 14.7  \\
    & + ROS  &     33.7 / 12.6     &  42.9 / 20.4 \\
    & + SN   & 49.3 / 20.5   &  62.3 / 34.2 \\ 
    
\midrule
\multirow{3}{*}{$\alpha = 1.5 $} 
    & Original  &  3.8 / 1.0 &   4.8 / 1.4    \\
    & + ROS  &  9.6 / 1.2 &  14.6 / 2.3    \\
    & + SN &   21.8 / 3.3  &   35.8 / 6.9  \\

\bottomrule
\end{tabular}
\centering
\end{table}

\section{Visualizations}\label{appendixsec:visuals}

We now visualize the predictions made by models before and after equipping with EdgeHead and illustrate their differences. As shown in Figure~\ref{fig:visual_1} and Figure~\ref{fig:visual_2}, the predictions made by models equipped with EdgeHead match the ground truth better on the closer surfaces to the ego vehicle. Without changing the size of the predicted boxes much, the proposed EdgeHead guides the models to focus more on the closer surfaces and the closest box corner by refining the location and rotation of the boxes.

\begin{figure*}[h]
\centering
\includegraphics[width=0.95\textwidth]{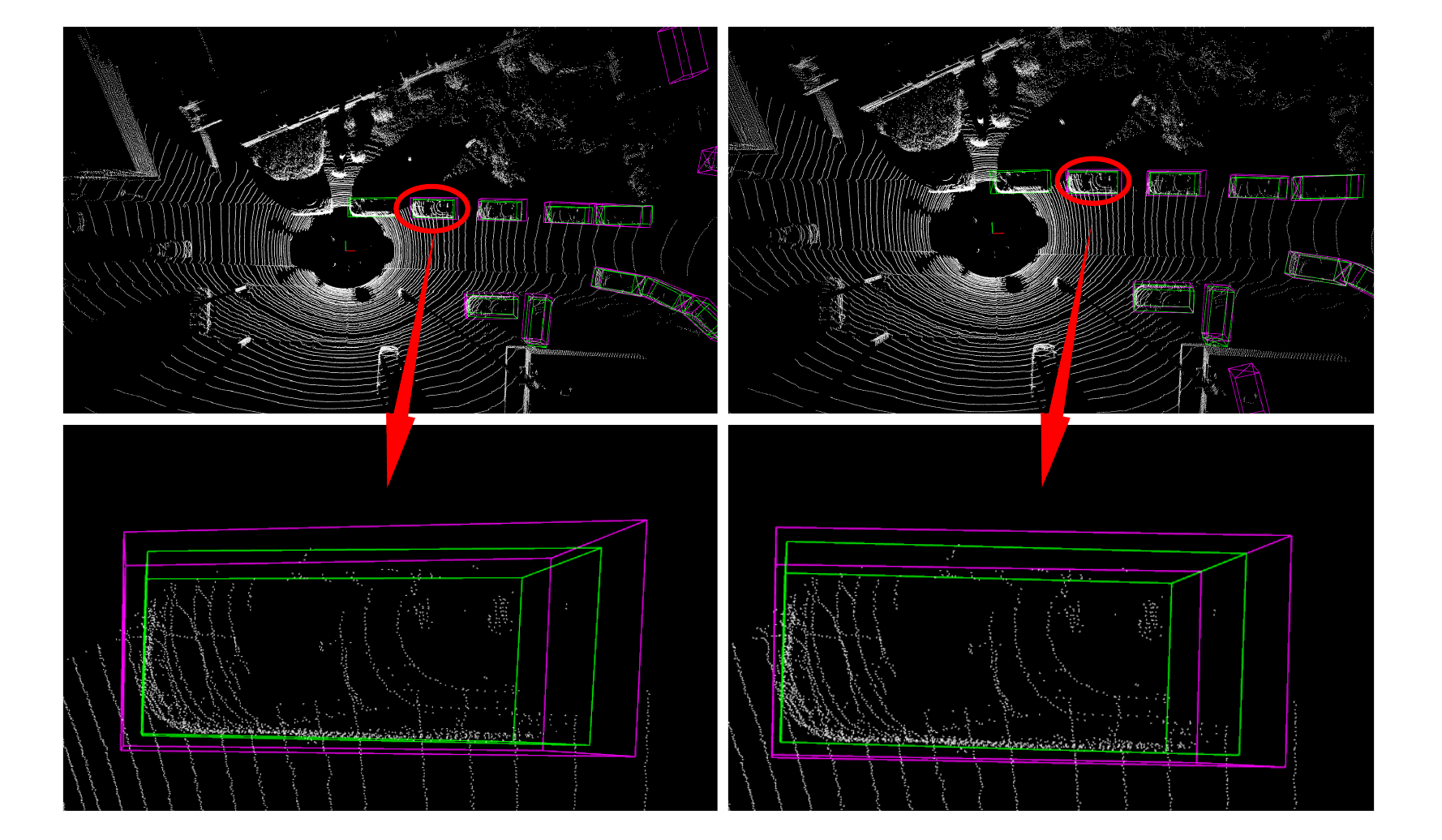}
\caption{\textbf{Visualizations of SECOND (left) and SECOND+ROS+EdgeHead (right) for the Waymo $\rightarrow$ KITTI task.} We use magenta and green boxes to show the predictions and ground truth, respectively.}
\label{fig:visual_1}
\end{figure*}

\begin{figure*}[h]
\centering
\includegraphics[width=0.95\textwidth]{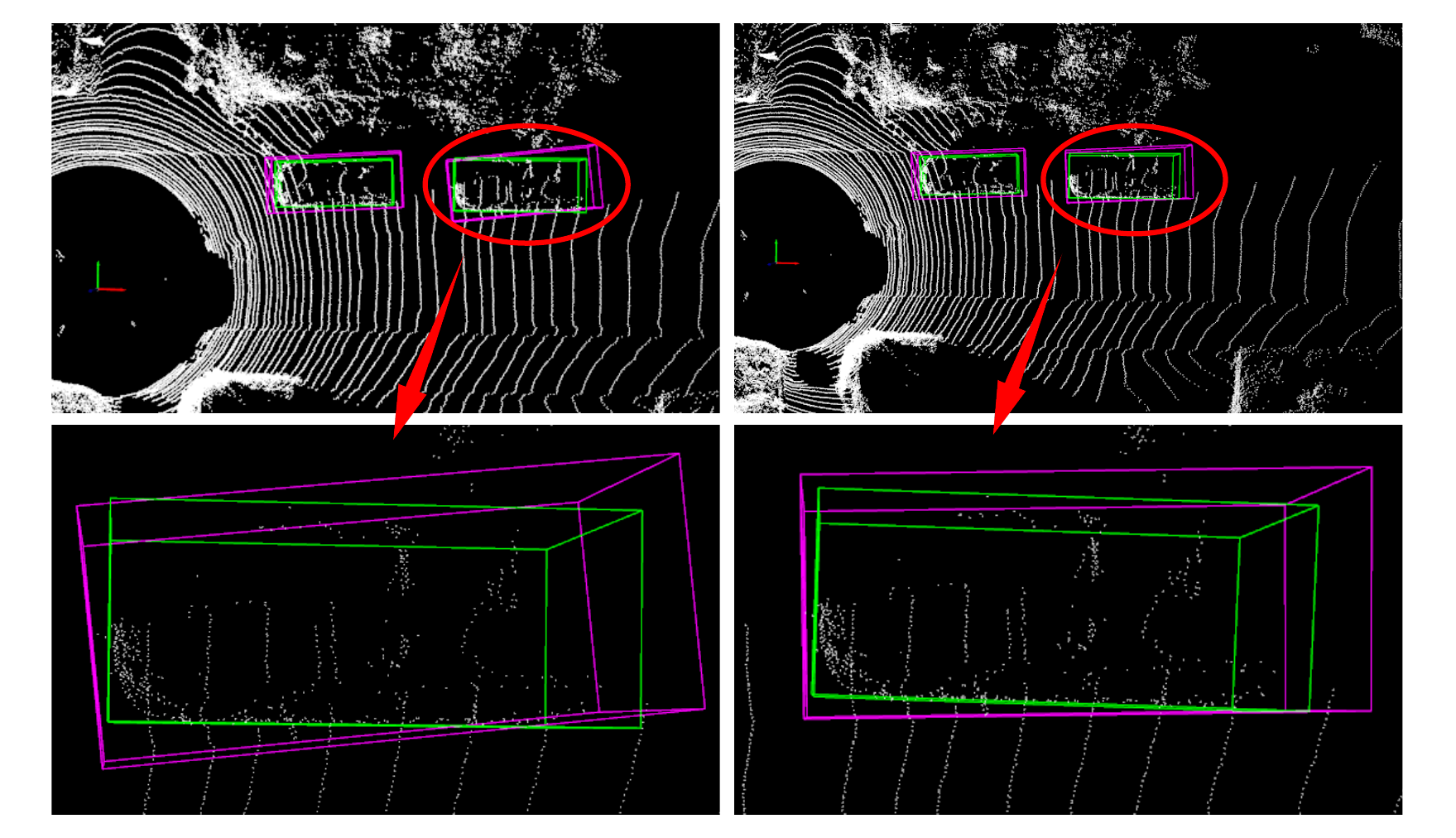}
\caption{\textbf{Visualizations of CenterPoint (left) and CenterPoint+EdgeHead (right) for the nuScenes $\rightarrow$ KITTI task.}}
\label{fig:visual_2}
\end{figure*}

\section{Setting of $\alpha$ in the new metrics}\label{appendixsec:alpha}

We finally analyze the different settings of $\alpha$ in the proposed CS-ABS and CS-BEV metrics. We set the value of $\alpha$ as 1 by default in the main paper, while we investigate the performance influence of different settings of $\alpha$ here. As shown in Table~\ref{tab:appendix_alpha}, when setting $\alpha$ to smaller values (\eg 0.5), it makes the penalty to the original BEV IoU smaller and thus provides closer CS-BEV results for the original SECOND model before and after equipping with EdgeHead. Conversely, setting $\alpha$ to larger values (\eg 1.5) could diminish the models' detection performance. Therefore, a more reasonable choice of $\alpha$ should be better in the range of $(0.5, 1.5)$ for practical use.

\end{document}